\documentclass{article}

\usepackage[nonatbib,preprint]{neurips_2022}

\usepackage[utf8]{inputenc} 
\usepackage[T1]{fontenc}    
\usepackage{hyperref}       
\usepackage{graphicx}
\usepackage{url}            
\usepackage{booktabs}       
\usepackage{amsfonts}       
\usepackage{nicefrac}       
\usepackage{microtype}      
\usepackage{xcolor}         
\usepackage{amsmath}
\usepackage{subfig}
\usepackage{cleveref}
\Crefname{figure}{Figure}{Figures}
\crefname{figure}{Figure}{Figures}
\Crefname{section}{Section}{Sections}
\crefname{section}{Section}{Sections}

\usepackage[algoruled,lined,linesnumbered,longend]{algorithm2e}

\newcommand{\nlive}{\ensuremath{n_\text{live}}}

\newcommand{\Z}{\ensuremath{\mathcal{Z}}\xspace}
\newcommand{\logZ}{\ensuremath{\log\Z}\xspace}
\newcommand{\like}{\ensuremath{\mathcal{L}}\xspace}
\newcommand{\post}{\ensuremath{\mathcal{P}}\xspace}
\newcommand{\prior}{\ensuremath{\Pi}\xspace}
\newcommand{\evidence}{\ensuremath{\mathcal{Z}}\xspace}

\newcommand{\nprior}{\ensuremath{n_\text{prior}}\xspace}
\newcommand{\anyg}[3]{\ensuremath{#1\mathopen{}\left(#2\,\rvert\, #3\right)\mathclose{}}\xspace}
\newcommand{\ofOrder}[1]{\ensuremath{\mathcal{O}\left(#1\right)}}

\newcommand{\PC}{\textsc{PolyChord}\xspace}

\newcommand{\W}{\ensuremath{\mathcal{W}}\xspace}
\newcommand{\Wsym}{\ensuremath{\W\mathrm{sym}}\xspace}
\newcommand{\Wskew}{\ensuremath{\W\mathrm{skew}}\xspace}

\title{Split personalities in Bayesian Neural Networks:\\ the case for full marginalisation}

\author{%
    David Yallup\thanks{Corresponding Author}\\
    Kavli Institute for Cosmology\\
    Cavendish Laboratory \\
    University of Cambridge\\
    \texttt{dy297@cam.ac.uk} \\
    \And    
    Will Handley\\
    Kavli Institute for Cosmology\\
    Cavendish Laboratory \\
    University of Cambridge\\
    \texttt{wh260@cam.ac.uk} \\
    \And    
    Mike Hobson \\
    Astrophysics Group\\
    Cavendish Laboratory \\
    University of Cambridge\\
    \texttt{mph@mrao.cam.ac.uk}\\
    \And    
    Anthony Lasenby \\
    Astrophysics Group\\
    Cavendish Laboratory \\
    University of Cambridge\\
    \texttt{a.n.lasenby@mrao.cam.ac.uk}\\
    \And    
    Pablo Lemos \\
    Department of Physics and Astronomy\\
    University of Sussex\\
    \texttt{p.lemos@sussex.ac.uk}\\ \\
}

\begin{document}

\maketitle

\begin{abstract}
    The true posterior distribution of a Bayesian neural network is massively multimodal. Whilst most of these modes are functionally equivalent, we demonstrate that there remains a level of real multimodality that manifests in even the simplest neural network setups. It is only by fully marginalising over all posterior modes, using appropriate Bayesian sampling tools, that we can capture the split personalities of the network. The ability of a network trained in this manner to reason between multiple candidate solutions dramatically improves the generalisability of the model, a feature we contend is not consistently captured by alternative approaches to the training of Bayesian neural networks. We provide a concise minimal example of this, which can provide lessons and a future path forward for correctly utilising the explainability and interpretability of Bayesian neural networks.
\end{abstract}

\section{Introduction}\label{sec:intro}

The emergence of the deep learning paradigm to tackle high dimensional inference problems has driven numerous striking advances in recent years. In most practical settings this is approached as a high dimensional Neural Network (NN) optimisation problem, with an extensive industry built up to attempt to make this reliably solvable. Bayesian Neural Networks (BNNs) are a parallel program to bring probabilistic reasoning over the same classes of models, realised through the application of Bayes theorem to the training procedure. A core attraction of performing Bayesian inference with NNs is that it provides a probabilistic interpretation of the action of a network, allowing clear expression of the increasingly important question: \emph{How certain are the predictions we make with ML?} Vital issues such as quantifying the potential for bias to occur in network predictions are expressed in a principled Bayesian framework; imbalances in input training data are fed back as reduced certainty in predictions for under-represented population members in the training sample~\cite{2021arXiv210604015N,maddox2019simple}.


Despite the attraction of a BNN framework, progress in this area has been marred by a looming issue: performing Bayesian inference over the vast parameter spaces a typical deep NN covers. The increased utility of a Bayesian calculation suffers a trade-off in the dimensionality of parameter space that the inference can be drawn over. When analytic approaches break down, as is the case in any real-world scenario, approximations are needed to make the Bayesian computations tractable~\cite{foong2020expressiveness}. Particular attention is generally paid to Bayesian approximations that are best suited to scaling to the very high dimensional space that maximum likelihood based approaches typically operate in. In general this results in an approximation for the posterior network probability distribution that holds under some potentially strict assumptions~\cite{2020arXiv200202405W}. In particular it is often assumed that the posterior over the free parameters of a network is suitably captured by a unimodal distribution (typically around the maximum a posteriori value). In this work we propose a minimal problem that directly challenges this viewpoint by demonstrating that even the simplest possible Neural Networks have a posterior distribution that is truly multimodal.

In the example we provide in this work, we present the case for focusing on a ``compromise free'' numerical marginalisation of the likelihood as a way to guarantee that the training procedure accounts for, and accurately reflects, the posterior multimodality. We perform this explicit marginalisation using numerical sampling tools designed to directly calculate the full marginal likelihood, or \emph{Evidence}. A suitable sampling tool for such a calculation should have the ability to handle integrals over likelihoods with potentially multiple strongly peaked structures, which in turn affords formal guarantees that the posterior distribution inferred from the samples accurately captures these structures. Some preliminary work in marginalising over NNs in this manner has taken place~\cite{bsr,javid2020compromisefree}, where in general a focus has been placed on the role of the evidence for Bayesian Model comparison between networks. Whilst this is an interesting avenue to build further on the work presented here, we make the case that by closely examining the split personalities of networks emerging from these calculations, a fundamental challenge to all probabilistic NN descriptions is exposed.

In this work the case for, and an example of, explicit marginalisation over NNs is presented. In \cref{sec:bayes} a brief overview of Bayesian Neural Networks is given, particularly covering the challenges facing numerical Bayesian methods. In \cref{sec:ex} we present a minimal challenge that exposes the truly split behaviour of a NN posterior in a clear fashion, as well as discussing the implications of this observation. Lastly in \cref{sec:conc} we provide some concluding remarks, and higher-level motivation for future work.

\section{Bayesian formulation of Neural Network training}\label{sec:bayes}

This work focuses on a subset of the myriad of modern NN realisations: Fully Connected Feedforward Neural Networks. These networks were some of the earliest proposed forms of Artificial Neural Network, and in spite of the rise of symmetry respecting transformation layers~\cite{DBLP:journals/corr/abs-2104-13478}, these \emph{Dense} (synonymous with fully connected) layers are still abundant in modern architectures. 

A dense layer is composed of a linear mapping defined by a weight matrix \W taking a vector of $M$ inputs to a vector of $N$ outputs, $\W : \mathbb{R}^M  \rightarrow \mathbb{R}^N$. The elements of this matrix, $w_{ij}$, connect every input element $i \in M$ to every output $j \in N$, a constant \emph{bias} vector, $b_j$, is also usually added. If the output nodes of a dense layer are internal (or hidden) a non-linear \emph{activation} function is applied to each node. Repeated action of $K$ (indexed in this work as a superscript $k\in\{1,\dots,K\}$)  parameterised linear maps with activation builds up a deep NN, mapping an input data vector, $x_i$, to a prediction vector, $y'_j$. A neural network would be considered deep if it has more than one hidden layer ($K\geq3$), in this work we examine \emph{shallow} networks ($K=2$).

In this study we focus on a simple binary classification network, predicting class probabilities of a categorical distribution likelihood. The network is then trained by comparing labelled data, $D=\{x,y\}$, to predictions, $y'$, made with a choice of the free parameters of the network, $\theta=\{w,b\}$. The comparison is realised through the joint likelihood for making a set of observations conditioned on a choice of $\theta$, $\anyg{\like}{D}{\theta}$. The Bayesian approach to training is then using Bayes theorem to invert the conditional arguments of this likelihood,
\begin{equation}
    \anyg{\post}{\theta}{D}=\frac{\anyg{\like}{D}{\theta} \prior(\theta)}{\evidence (D)}\,,
\end{equation}
where in addition to the likelihood, the following distributions are introduced: the prior over the network parameters \prior, the posterior distribution of the network parameters \post and the evidence for the network \evidence. The evidence is the marginal likelihood, found by marginalising over all the network parameters,
\begin{equation}
    \evidence(D) = \int \anyg{\like}{D}{\theta} \prior(\theta) d\theta\,.
\end{equation}
A trained network in this sense is the posterior distribution of the weights and biases given an input set of data points. A sample from this posterior distribution can then be used to predict a sample of predicted labels for a new unseen set of data points, referred to as the model prediction through the remainder of this article. The role of the Bayesian evidence in tasks such as model selection is long established~\cite{mackay}, and has been demonstrated as a tool for tuning hyperparameters such as number of nodes in a hidden layer of a network~\cite{bsr,javid2020compromisefree}. What is less clearly established thus far in the literature are the implications of obtaining a convergent evidence calculation in relation to the resulting network posterior distribution.

\subsection{Priors for Dense Neural Networks}\label{sec:prior}
Since the early days of NNs it has been understood that there are many functionally equivalent, or \emph{degenerate}, choices of $\theta$ for any given network architecture~\cite{10.1162/neco.1994.6.3.543}. For any given dense layer outputting $N$ nodes, arguments arising from combinatoric reordering of nodes and potential compensating translations across activation functions, lead to $2^N N!$ degenerate choices of weight for every posterior mode~\cite{bishop}. This presents a strict challenge for any Bayesian method. Either you sample the posterior and find at least $2^N N!$ degenerate modes per layer (noting this is clearly impossible due to the combinatorial growth of this factor and the typical value of $N$ in modern NNs), or you don't and hence have only explored some undefined patch of the total parameter space. In order to isolate and examine this issue in detail we employ a novel definition for the solution space a dense NN occupies. We propose that a constrained form of weight prior can be employed that retains the flexibility of a dense layer, without admitting functionally equivalent solutions.


We initially restrict our analysis to mappings with $M=N$, \emph{i.e.} only admitting square weight matrices.
A square weight matrix can be trivially written as a sum of symmetric and skew symmetric pieces,
\begin{equation}
  \W = \frac{1}{2} (\W + \W^T) + \frac{1}{2} (\W - \W^T) = \Wsym + \Wskew \,.
\end{equation}
We can impose a condition that \Wsym is positive definite. This implicitly removes $N!$ permutations by requiring that the weight matrix be pivot free. Further to this requiring positive definiteness imposes that the scaling part of the linear map cannot flip sign, giving consistently oriented scaling transformations. The remaining degrees of freedom in \Wskew can also be constrained, \Wskew forms the Lie algebra of $SO(N)$, with the structure constant being the totally anti-symmetric tensor $\epsilon$. By convention we can take the normal ordering, $\epsilon_{1,\dots,N} = 1$, which in practice can be enforced by appropriate sign choices in \Wskew. 

For simplicity in this work we use a uniform prior over weights and biases in the range $(-5,5)$, rejecting any samples where \Wsym is not positive definite, and \Wskew is not consistently oriented. It would perhaps be more familiar to choose a Gaussian prior, or equivalently L2 regularisation, to promote sparsity. This is less important as we are not operating in the over-parameterized regime with the examples we study, although implications of non flat priors are briefly discussed alongside our results. Starting from any typical symmetric prior and rejecting to a particular subset of mappings is useful for visualising this proposal on the concise problem in this work, however these constraints can be more efficiently encoded as a prior that has an appropriate support. The $N\times N$ matrix can be effectively split into a lower triangle (including diagonal), $L$, and upper triangle, $U$. \Wsym can then be generated as $\Wsym = L\cdot L^T$ with a prior over $L$ to reflect the desired sparsity promotion (typically employed in the form of regularisation terms for optimized NN approaches). The upper triangle provides the skew piece as, $\Wskew = U -U^T$, with appropriate signed priors over $U$ to respect a choice of structure constant for $SO(N)$. 

The generalisation of this to cases where $M\neq N$ is not trivial, particularly to construct in the rejection sampling format used in this work.
A potential construction of a prior with this support could come by considering the reverse of a singular value decomposition. It is typical for such a decomposition of \W to implicitly perform the required pivoting to return a sorted and positive set of singular values, as well as orthogonal rotations in the $M$ and $N$ planes. To remove spurious degeneracy that emerges due to the flexibility of these rotations a similar construction to  the one motivated for appropriate \Wskew in square matrices is likely necessary. Additionally we note that the symmetry respecting equivariant layers being increasingly deployed in modern NN applications are naturally better suited than dense layers to the marginalisation we attempt in this work~\cite{equiv}. It is quite natural to incorporate such layers into the picture we build here.




\subsection{Tools for Neural Network marginalisation}\label{sec:NSforNN}
In \cref{sec:prior} we considered the potentially troublesome sampling problem a dense NN layer poses. In order to sample these spaces we can consider which numerical Bayesian tools will be sufficient for adequately exploring the parameter space.

Limited exploration is acceptable if every mode is functionally equivalent, however if there is genuine multimodality to the solution, there is no guarantee that the sampled posterior will be representative of the true posterior unless these degeneracies can be resolved before sampling. Bayesian methods that arise from an optimized point estimate (Variational Inference being a common example of this~\cite{tomczak2020efficient,rudnertractable,2018arXiv181003958W}), by construction cannot reflect any split behaviour. Sampling methods such as Hamiltonian Monte Carlo do give formal guarantees to sample a representation of the full density~\cite{cobb2019semi,cobb2021scaling}, however in practice it is well known that multimodal targets are a common failure mode for Markov Chain approaches. Ensemble approaches (either of Markov chains, or optimized point estimate solutions) are a potential method to extend the previously mentioned tools to work with multimodal targets. Whilst there is promise in these approaches, in general they lack formal probabilistic guarantees to obtain the true posterior~\cite{2020arXiv200110995W}. In practice then, it is often considered sufficient to either implicitly (or explicitly) assume the posterior distribution is unimodal. 

A correctly weighted ensemble of unimodal methods would adequately define the correct posterior, however the challenge is in how to construct this weighting. In this work we contend that an effective approach can be found by employing Nested Sampling~\cite{Skilling:2006gxv} (NS). NS is a sampling algorithm uniquely designed to target numerical estimation of the evidence integral. It is often bracketed with other Markov Chain sampling algorithms, but its ability to navigate strongly peaked multimodal distributions with little to no tuning makes it uniquely able to perform a full marginalisation of a Neural Network, in a manner necessary for this work. As a full numerical integration algorithm, the dimensionality of space that the sampling can cover is typically limited to \ofOrder{1000} parameters. Whilst this is much smaller than desired for full networks, probabilistic techniques with similar dimensionality limitations, such as Gaussian Processes, have been successfully embedded in scalable frameworks~\cite{DBLP:journals/corr/WilsonHSX15,2017arXiv171100165L}.

In this work we use the \PC implementation of NS~\cite{Handley:2015vkr}. The resolution of the algorithm is defined by the number of live points used and is set to $\nlive=1000$, the number of initial prior samples is also boosted to ensure exploration of the space and set to $\nprior=\nlive\times 10$. All other hyperparameters are set to the defaults. The chain files are analysed using the \texttt{anesthetic} package~\cite{anesthetic}, which is used to create the corner plots. The NNs were built on the \texttt{flax}~\cite{flax2020github} NN library within the \texttt{jax} ecosystem~\cite{jax2018github}.

\section{A concise example of multimodality}\label{sec:ex}

\begin{figure}
  \centering
  \subfloat[The minimal Neural Network used in this example.\label{fig:nn}]{\includegraphics[width=0.6\textwidth]{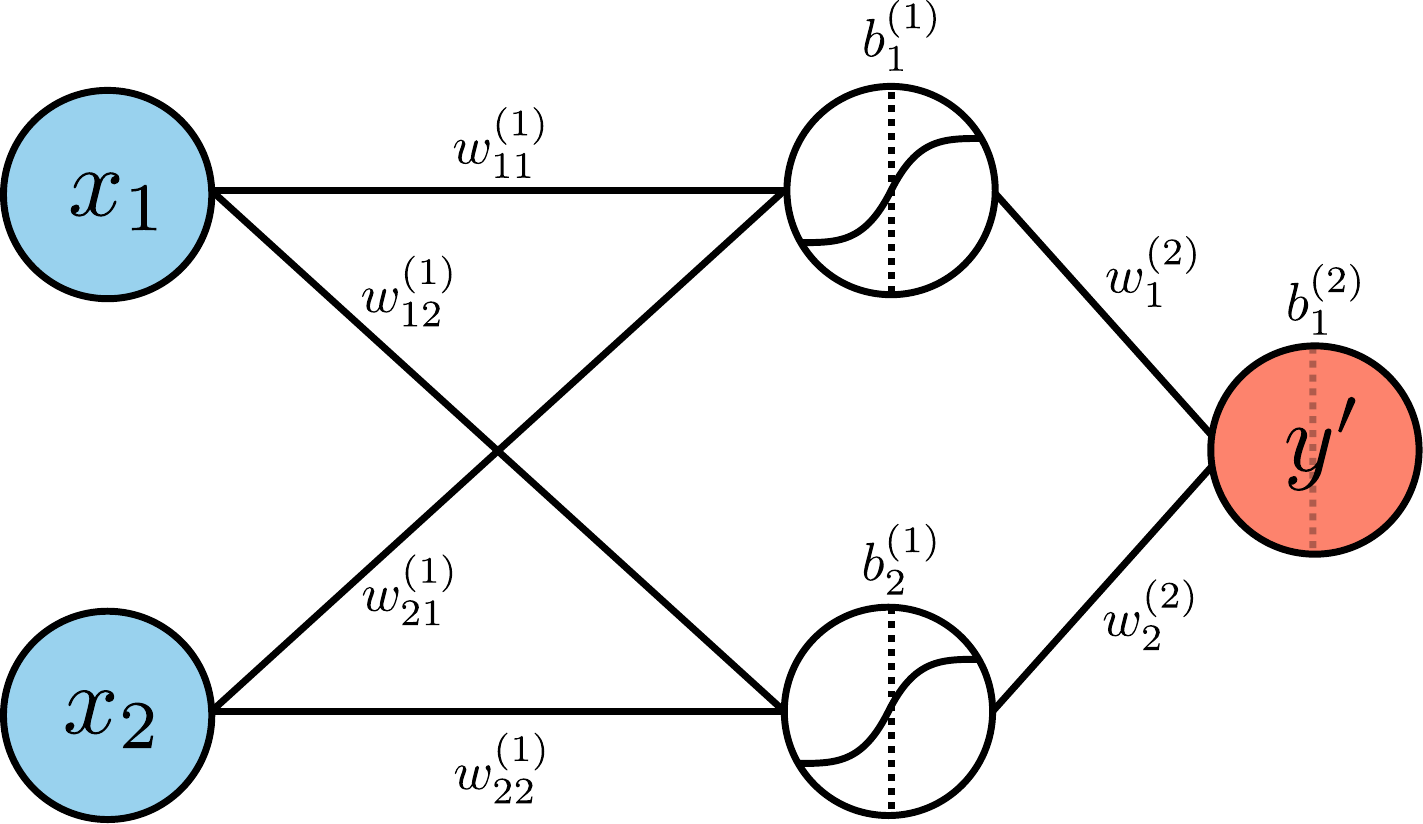}} \\
  \subfloat[The target training data, and trained model on the noisy XOR problem.\label{fig:target}]{\includegraphics[width=1.0\textwidth]{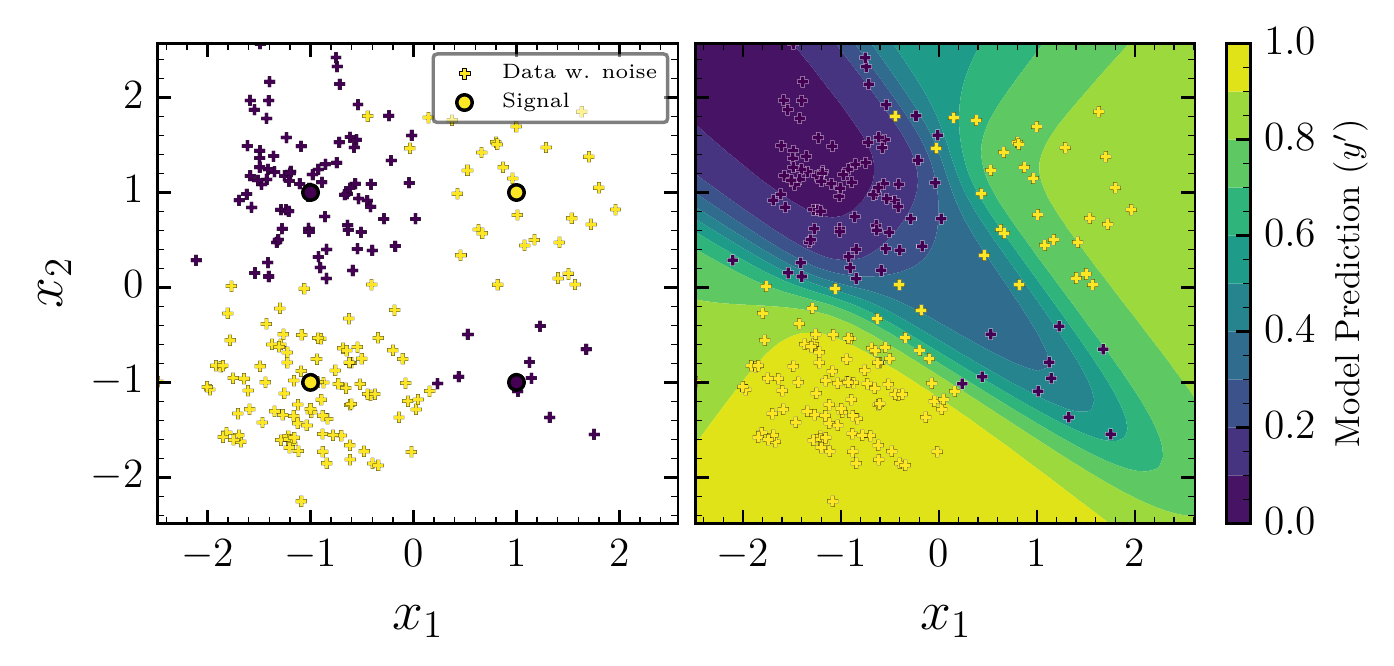}}
  \caption{The chosen network architecture (\cref{fig:nn}) and target function to fit in this challenge (\cref{fig:target}). The left panel of \cref{fig:target} shows the initial cardinal signal points, with noisy uneven sampled data along side this. The right panel of the same Figure retains the same noisy training data overlaid on a prediction from the trained NN across the whole parameter space.}
\end{figure}

In this section we propose a minimal classification problem, that can be fit with a minimal NN. We choose to examine a modified noisy version of an XOR logic gate problem. Samples forming the training set are drawn from cardinal signal points $(x_1,x_2) =\{(1,1), (-1,-1), (1,-1), (-1,1) \}$, with diagonal points being assigned a class $y=1$ and anti-diagonal points classified as $y=0$. The samples have two sources of noise added to remove any actual symmetry from the problem (with the intention of isolating degenerate symmetries in the weight space): samples are drawn unevenly from each cardinal point with $\{30, 100, 10, 80\}$ draws respectively, and to each sampled point Gaussian noise with a mean of 0 and a variance of 0.5 is added. Arbitrary test datasets can be made either by repeating the sampling with different cardinal populations or alternative noise seeds. A minimal neural network is chosen to fit this composed of a single hidden layer with $N=2$ nodes, passed through sigmoid activations with an added bias vector. Such a set up is typically identified as a logistic regression problem. This introduces a network with 9 free parameters, illustrated in \cref{fig:nn}. This network is trained with a standard binary cross entropy loss function\footnote{~{The binary cross entropy being the log of a categorical likelihood with two classes}}. Both the proposed problem and trained network solution are shown in \cref{fig:target}.


\begin{figure}
  \centering
  \subfloat[\label{fig:posta}]{\includegraphics[height=.4\textheight]{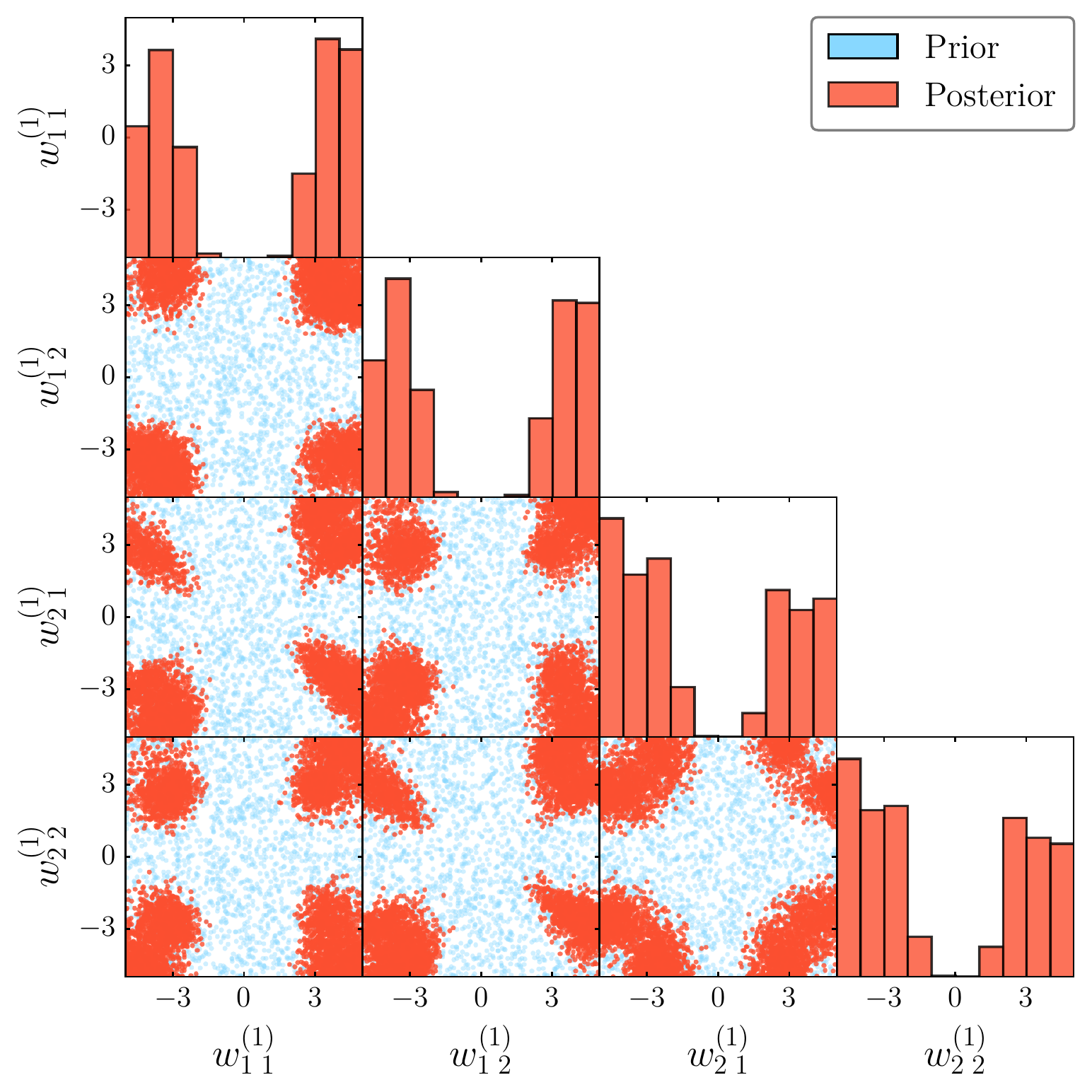}} \\
  \subfloat[\label{fig:postb}]{\includegraphics[height=.4\textheight]{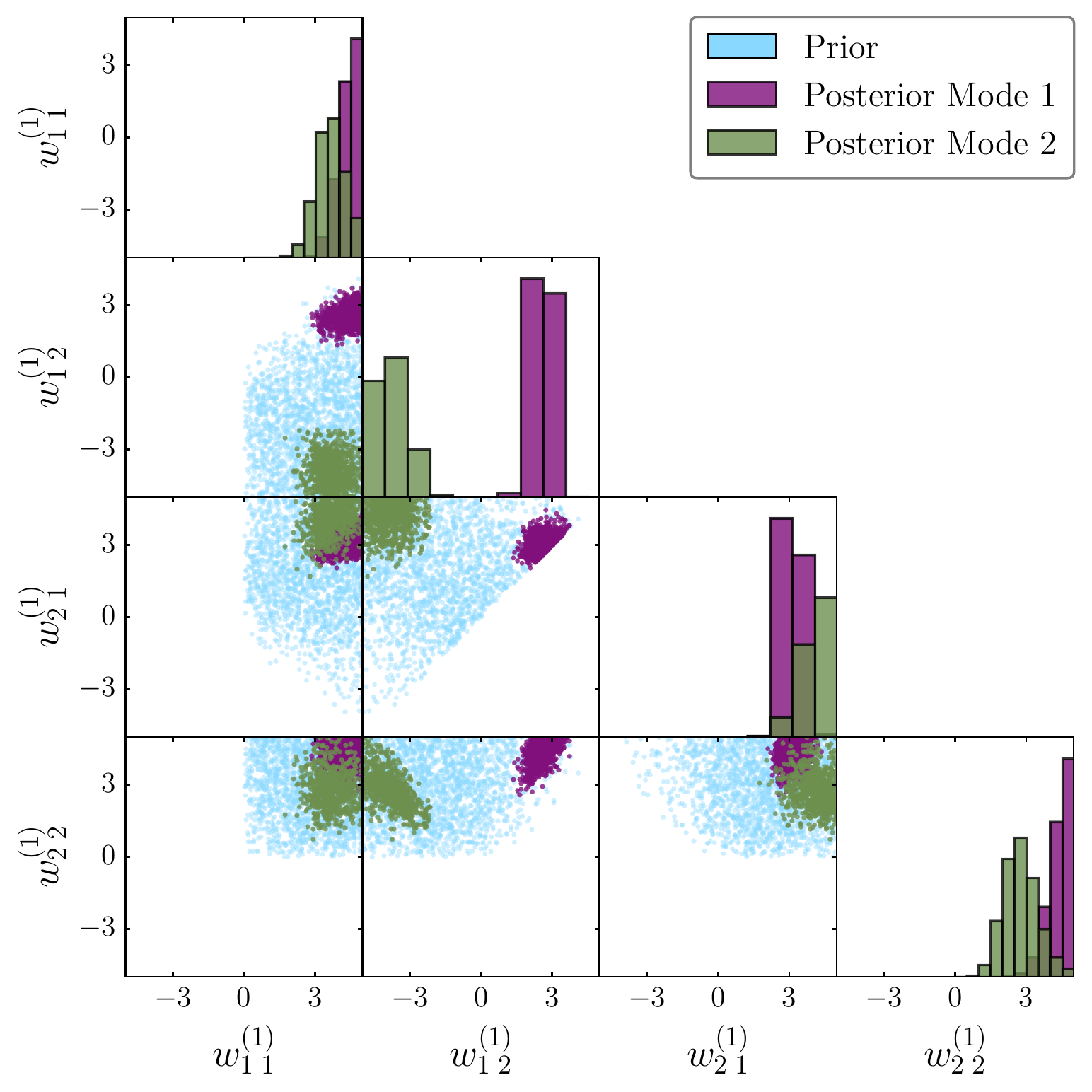}}
  \caption{Posterior samples from two choices of prior for a dense layer, where the parameters shown are only the weight matrix parameters connecting the input data to the hidden nodes. A Uniform distribution from (-5,5) forms the base weight prior of both figures, with \cref{fig:postb} rejecting samples not following the recipe given in \cref{sec:prior}. The distinct modes in \cref{fig:postb} have been separated and coloured differently, the sum of these densities gives a functionally equivalent network to the red posterior in \cref{fig:posta}.\label{fig:posterior}}
\end{figure}

The network predictions shown in \cref{fig:target} are built by taking the mean prediction of the binary classifier $y'$ over 1000 posterior samples of $\theta$. These samples are drawn from the posterior calculated as the result of using Nested Sampling to fully marginalise the network likelihood, as detailed in \cref{sec:NSforNN}. In the model prediction figures throughout this work we display the mean of the vector of predicted $y'$ over a regular grid of input data $(x_1,x_2)$ with spacing 0.1 covering the full set of training points. We can further interrogate the structure of these solutions by examining how the posterior samples of network parameters, $\theta$, are distributed. The corner plots of these posteriors are shown in \cref{fig:posterior}. For simplicity only the parameters of the weight matrix connecting the inputs to the hidden nodes are shown. This figure displays posterior samples for networks trained with two different choices of prior, with an important additional note that both choices of prior give functionally equivalent model predictions (\emph{e.g.} the right panel in \cref{fig:target}). The Nested Sampling algorithm realised in \PC keeps track of the number of posterior modes it finds during its evolution. We define a prominent node as being any mode that has a local evidence estimate greater than a factor of $10^{-3}$ smaller than the mode with the highest local evidence. This definition implies that we only select modes that contribute on average at least 1 sample to the 1000 posterior samples used to infer the figures in this work. By this definition the full prior finds 14 distinct prominent modes, whereas the proposed prior constraint yields only 2 prominent posterior modes.

\begin{figure}
  \centering
  \includegraphics[width=1.0\textwidth]{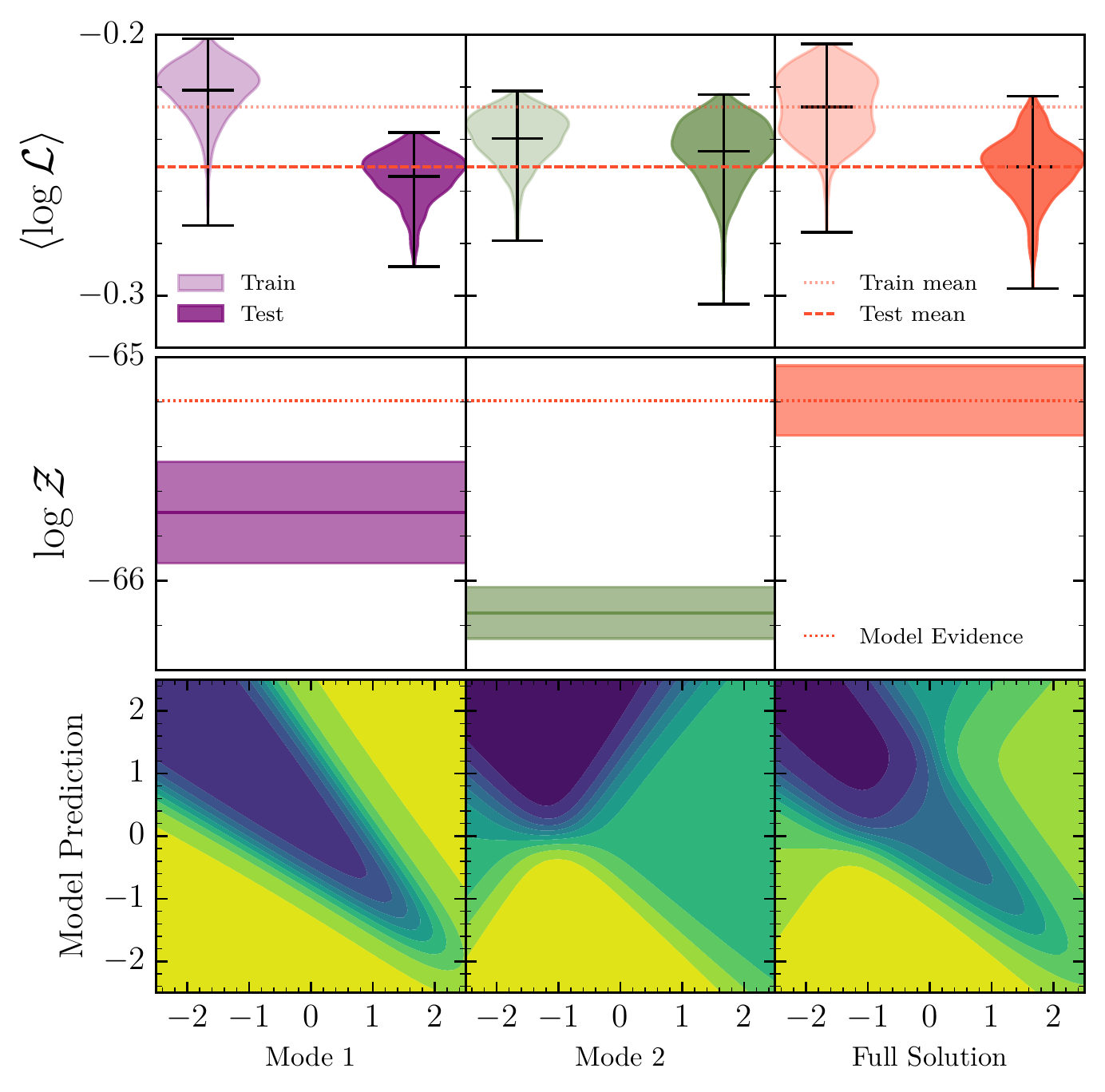}
  \caption{Comparison diagnostics for the two posterior modes found when using the constrained prior, as well as the corresponding full solution. The $\langle \log \like \rangle$ and \logZ diagnostics for each solution are colour coded corresponding to the colours used in \cref{fig:posterior}, with the Model Prediction diagnostic corresponding to colouring and scale used in \cref{fig:target}. \label{fig:diagnostic}}
\end{figure}

The proposed form of prior for dense layers greatly simplified the composition of the target function, from being a weighted superposition of 14 modes to only 2, with no apparent change in the quality of the predictions. This can be further investigated by now examining the individual modes found against the full solution already presented, this is shown in \cref{fig:diagnostic}. The rows of this figure display three comparison diagnostics, with the columns replicating the diagnostics for each mode as well as the full solution. The top row of panels display the average log likelihood, $\langle \log \like \rangle$, across the training and test dataset. A test dataset has been formed for this purpose by taking ten repeated samples of the same cardinal point proportions as in the training set, but with a different Gaussian noise seed for each set. The violin density estimate is again derived from an ensemble of predictions made using 1000 posterior samples (from the individual modes and the full posterior respectively), with the error bars in black showing the mean $\langle \log \like \rangle$ from these posterior set and extending to the most extreme values in the ensemble. Overlaid across all three panels are the mean values from the full solution. The maximum a posteriori mode (containing as well the maximum likelihood solution in this set), is Mode 1. Whilst this mode gives the best performing (highest average $\langle \log \like \rangle$) solution on the training set it poorly generalises to the test set, under-performing the full solution on average. Mode 2 gives a more robust and generalisable solution, outperforming the full solution on the test set on average however it less successfully captured the training set. It is worth noting that this kind of ``local minima'' in the loss landscape is deliberately suppressed in nearly all NN training schemes. The full solution, as an evidence weighted superposition of the two modes combines the best features of both. The second row of panels display the log evidence of each solution, $\logZ$. The log evidence of each mode effectively defines the full solution, with the ratio of evidences giving the ratio of contribution of each mode to the full solution. The bottom row of panels show the predictions of each solution across the input data space, using the same scale and format as the corresponding panel in \cref{fig:target}.

\subsection{Discussion}

The noisy XOR gate problem demonstrated in this work provides a clear illustration of multimodal behaviour in NN posteriors, present in even the most minimal network. This poses challenges both for future work using the principled marginalisation approach we proposed in this study, and more generally for robust inference with any NN. There are however some marked differences between the stylised challenge presented here and the generally accepted standard practices for NN inference, here we consider how some of these standard practices would impact the patterns demonstrated.

A notable feature that only emerges from the full solution as demonstrated in \cref{fig:diagnostic}, is the complexity of the model achieved. It is not possible to represent the decision boundary that the full multimodal solution constructs with any single point estimate network weights given the fixed finite network size. Correctly posterior weighting two candidate models gives an inferred decision boundary that would only be achievable with a point estimate solution for a higher dimensional network. It is not uncommon for people to consider Bayesian inference as simply giving an uncertainty on a prediction, but in this case it is doing something grander, allowing construction of higher dimension solutions by a mixture of models. A larger model with standard regularization terms (or equivalently non-flat priors) could be proposed as a way of constructing a more unimodal solution. Reshaping the prior will influence the evidence calculation, and hence the ratio of local evidences between modes, but there is little motivation to think that this will consistently decide in favour of a particular solution. Furthermore we contend that increasing the dimensionality of the model will likely make the potential for competitive models at scale worse.

Reading further into the diagnostic plot, the maximum likelihood solution (Mode 1) is demonstrated to have worse generalisability than the full solution. By allowing a weighted sum of two distinct models to contribute to our full solution, a network that is more robust to noise is constructed. Whilst this in and of itself is a useful trait for any network, there is a secondary fact about principled Bayesian inference on display here, robust performance in low data/high noise environments. Taking the typical argument that Neural Networks are data hungry, this challenge could be refuted by requiring a much larger set of data samples for training, implicitly averaging over more samples of data noise at training time. Whilst this would resolve much of the challenge in this case, this is an ad-hoc requirement for applicability of NNs and furthermore an inadequate defence if we want to make sensitive statements in the massive model and data limits employed in cutting edge large model NN inference~\cite{DBLP:journals/corr/abs-2110-09485}.

Lastly, a potential refutation of this challenge is to invoke some kind of ensemble based method as a vehicle to capture a set of candidate alternative models. The correct weighting of the two proposed solutions in our fully marginalised approach is the result of a convergent calculation of the exact volume under the likelihood surface over the prior. Any method that does not guarantee that the ratio of densities is correct will have a vanishingly small chance of getting this ratio correct, the rate of attraction of a gradient boosted path (whether that be via an optimizer or an HMC chain) will not necessarily be proportional to the posterior mass of a mode.


\section{Conclusion}\label{sec:conc}
In this work we presented a concise example of multimodal posteriors in Neural Networks and the case for capturing them with numerical marginalisation. By reducing the Neural Network to its most minimal form, and choosing a minimal simple learning problem, a paradox for inference using Neural Networks was posed. In this paradox, a learning scheme that focuses on network parameters that maximise the likelihood on the training data is at odds with the aim of generalisability and explainability of the network. It is only by both abandoning point estimate optimisation based approaches, and adopting a sampling scheme that can handle multimodal posterior distributions, that the most robust solution on this problem can be obtained. Two particularly compelling features of the solution a full numerical marginalisation can uniquely obtain were highlighted; the ability to reason between multiple candidate models and reason even when there is limited data with sizeable noise. Both of these are compelling features that show characteristics of what could be called intelligence. 

The current prevailing wind of Machine Learning research is following the direction of increasingly large models and large datasets. There are clear advantages, and some incredibly compelling results, emerging from this work at large scale. The orthogonal viewpoint we presented in this work offers features that can complement the striking successes of modern Machine Learning at scale. A more principled understanding, offered by a marginal likelihood based approach, of some parameters in a larger model could be a path to bring together principled probabilistic models with inference at massive scale.

\begin{ack}

    DY, WH and ANL were supported by STFC grant ST/T001054/1. DY \& WH were also supported by a Royal Society University Research Fellowship and enhancement award.

    \texttt{PolyChord} is licensed to \href{https://raw.githubusercontent.com/PolyChord/PolyChordLite/master/LICENCE}{PolyChord Ltd.}, \texttt{PolyChordLite} is free for academic use. 

This work was performed using resources provided by the Cambridge Service for Data Driven Discovery (CSD3) operated by the University of Cambridge Research Computing Service (\url{www.csd3.cam.ac.uk}), provided by Dell EMC and Intel using Tier-2 funding from the Engineering and Physical Sciences Research Council (capital grant EP/T022159/1), and DiRAC funding from the Science and Technology Facilities Council (\url{www.dirac.ac.uk}).


\end{ack}



\bibliography{references}

\end{document}